\documentclass{article}

\usepackage{arxiv}

\usepackage[utf8]{inputenc} 
\usepackage[T1]{fontenc}    
\usepackage{hyperref}       
\usepackage{url}            
\usepackage{booktabs}       
\usepackage{amsfonts}       
\usepackage{nicefrac}       
\usepackage{microtype}      
\usepackage{graphicx}
\usepackage{doi}
\usepackage{multicol}
\usepackage{amsmath}
\usepackage{framed,multirow}

\title{UAV Images Dataset for Moving Object Detection from Moving Cameras}

\author{ \href{https://orcid.org/0000-0001-8119-2873}{\includegraphics[scale=0.06]{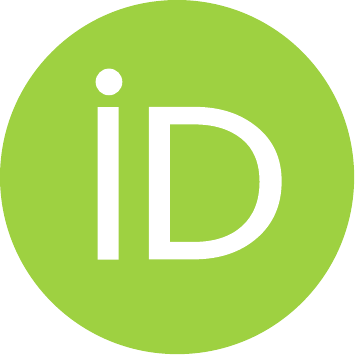}\hspace{1mm}İbrahim Delibaşoğlu} \\
	Department of Software Engineering\\
	Sakarya University\\
	Sakarya, Turkey\\
	\texttt{ibrahimdelibasoglu@sakarya.edu.tr} \\
}

\hypersetup{
pdftitle={A template for the arxiv style},
pdfsubject={UAV dataset, motion detection, moving object detection},
pdfauthor={Delibasoglu Ibrahim},
pdfkeywords={},
}

\begin{document}
\maketitle

\begin{abstract}

This paper presents a new high resolution aerial images dataset in which moving objects are labelled manually. It aims to contribute to the evaluation of the moving object detection methods for moving cameras. The problem of recognizing moving objects from aerial images is one of the important issues in computer vision. The biggest problem in the images taken by UAV is that the background is constantly variable due to camera movement. There are various datasets in the literature in which proposed methods for motion detection are evaluated. Prepared dataset consists of challenging images containing small targets compared to other datasets. Two methods in the literature have been tested for the prepared dataset. In addition, a simpler method compared to these methods has been proposed for moving object object in this paper.

\end{abstract}


\section{Introduction}
\label{sec:intro}

Unmanned aerial vehicles (UAV) and drones are nowadays accessible to everyone, and are used in civilian and military fields. Considering security applications, it is widely used in applications such as surveillance, target detection and tracking. The biggest problem in images obtained from such freely moving vehicles is that the background is always moving due to camera movement. The reasons such as the low resolution of the images obtained or the large distance to the target also make object detection and classification processes difficult. For this reason, it may be more convenient to apply motion detection as well as object detection in images obtained through UAV. Motion detection is a computer vision problem that has been extensively studied in the literature. Publicly available datasets used in motion detection are VIVID \cite{collins2005open} and Change Detection (CDNET-2014) \cite {wang2014cdnet}. In the CDNET dataset, only the PTZ sequence images were obtained from the moving camera. However, these images do not contain free motion as in the images obtained through UAV. The VIVID dataset consists of aerial images containing moving vehicles. In this study, a new dataset to be used in the problem of detecting moving objects from a moving camera is publicly shared with the researchers. The most important difference of the prepared dataset compared to the VIVID dataset is that it has high resolution (1920x1080) and contains more challenging small targets. The dataset called PESMOD (PExels Small Moving Object Detection) was created from videos selected from the Pexels website and moving objects was manually labelled. 

Also a method is proposed to detect moving object in this paper, and it is evaluated with two state-of-the-art methods in the literature. In section \ref{sec:relatedwork}, the related work of motion detection in the literature is introduced. The details of the dataset are represented in section \ref{sec:dataset}, and in the section \ref{sec:proposedMethod}, the details of the proposed method are explained. Finally, in sections \ref{sec:results} and \ref{sec:conclusion}, performance comparison is examined and the conclusion is discussed.

\section{Related work}
\label{sec:relatedwork}

Moving object detection is an issue that has been actively studied by computer vision researchers. Considering the motion detection studies in the literature; It is seen that methods based on the subtraction of consecutive frames, background modeling and dense optical flow are used. Although the method of subtraction consecutive frames works fast and can adapt quickly to background changes, the performance rate is very low \cite{collins2000system, zhao2011study}. In the literature, it is striking that successful results are obtained with background modeling methods, which are mostly created using certain previous frames. \cite{bouwmans2014traditional}. To build a background model, statistical methods \cite{moo2013detection, zivkovic2004improved, zivkovic2006efficient}, classical image processing techniques \cite{allebosch2015efic}, and neural networks \cite{de2017wisardrp} have been used in different studies. In addition to these, CNN (convolutional neural network) based methods have also been preferred, which have shown good performances in many image processing problems. Deep learning architectures that estimates pixel-based optical flow vectors have been used in the problem of moving target detection \cite{huang2018optical}. FlowNet2 \cite{ilg2017flownet} that is one of the widely used architecture in optical flow computing is used to estimate flow vectors, and background model is constructed in the
form of optical flow in the study \cite{huang2018optical}. The most important disadvantage of deep learning methods is that the computational cost is extremely high, especially in high resolution images.

Image alignment techniques are used to compensate camera movement (also called global motion), which is the most important problem in moving cameras. The principle idea is that the previous frame or background model is firstly warped to current frame with affine or perspective transformation matrix and then frame differencing or background subtraction is performed. In this study, performance comparison of two methods in the literature, MCD \cite{moo2013detection} and SCBU \cite{yun2017scene}, is performed on the prepared dataset.

In MCD and SCBU methods, Lucas Kanade method (KLT)\cite{tomasi1991detection} is used to track grid-based selected points for consecutive frames and RANSAC method\cite{fischler1981random} is used to calculate homography matrix representing camera motion. Then, background model is warped to current frame before updating background model and applying background subtraction. In MCD method, dual mode background model is constructed with mean, variance and age of each pixel. Mixing neighbouring technique is proposed to compensate the camera motion and it reduces the image alignment errors. In SCBU, a scene conditional background model updating method is proposed to handle dynamic scenes. It tries to build a clear background model without contamination of the foreground and foreground likelihood map is used to extract foreground regions by applying high and low threshold values. Our method applies same technique to compensate global motion. We apply neigborhood difference in background subtraction and the main contribution of the proposed method is that we apply weights acquired from dense optical flow vectors during background subtraction.

\section{Dataset}
\label{sec:dataset}

\begin{figure}[h]
\centering
\includegraphics[width=.8\textwidth]{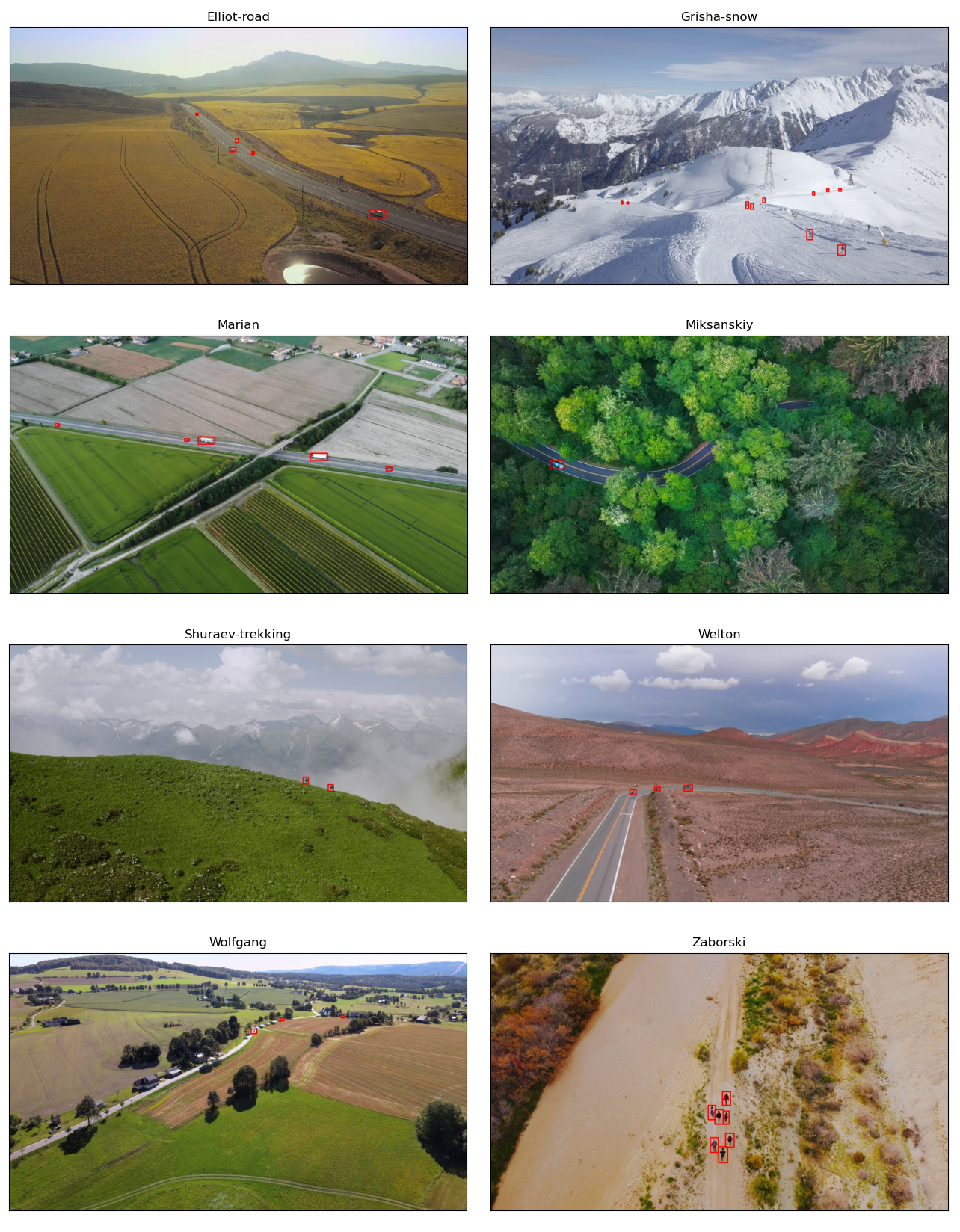}
\caption{Example frames from each sequence in the dataset}
\label{fig:dataset}
\end{figure}

While preparing the dataset, videos with small targets were selected from drone images that are open to public access from the Pexels website. Moving objects are labeled as a bounding box containing the object. The most important difference compared to VIVID and CDNET datasets is that it consists of high resolution images containing moving objects that are more difficult to detect (due to small sizes). The dataset consists of 8 different subset images. Only in one sequence called \emph{Grisha-snow}, the camera is fixed, but moving objects are very far away and difficult to detect in this sequence. Moving objects in images consist of people and different type of vehicles. Dataset contains totaly 4107 frames and 13834 labeled bounding box for moving targets. Details of each sequence are represented in the Table \ref{tab:datasetDetails}, and sample images are shown in the Figure \ref{fig:dataset}.

\begin{table}[h]
\centering
\caption{The details of PESMOD dataset}
\begin{tabular}{|c|c|c|}
\hline
\textbf{\begin{tabular}[c]{@{}c@{}}Sequence \\ name\end{tabular}} & \textbf{\begin{tabular}[c]{@{}c@{}}Number of \\ frames\end{tabular}} & \textbf{\begin{tabular}[c]{@{}c@{}}Number of \\ moving objects\end{tabular}} \\ \hline
\textit{Pexels-Elliot-road}                                       & 664                                                                  & 3416                                                                         \\ \hline
\textit{Pexels-Miksanskiy}                                        & 729                                                                  & 189                                                                          \\ \hline
\textit{Pexels-Shuraev-trekking}                                  & 400                                                                  & 800                                                                          \\ \hline
\textit{Pexels-Welton}                                            & 470                                                                  & 1129                                                                         \\ \hline
\textit{Pexels-Marian}                                            & 622                                                                  & 2791                                                                         \\ \hline
\textit{Pexels-Grisha-snow}                                       & 115                                                                  & 1150                                                                         \\ \hline
\textit{Pexels-zaborski}                                          & 582                                                                  & 3290                                                                         \\ \hline
\textit{Pexels-Wolfgang}                                          & 525                                                                  & 1069                                                                         \\ \hline
\end{tabular}
\label{tab:datasetDetails}
\end{table}

\section{Proposed method}
\label{sec:proposedMethod}

\begin{figure}[h]
\centering
\includegraphics[width=.6\textwidth]{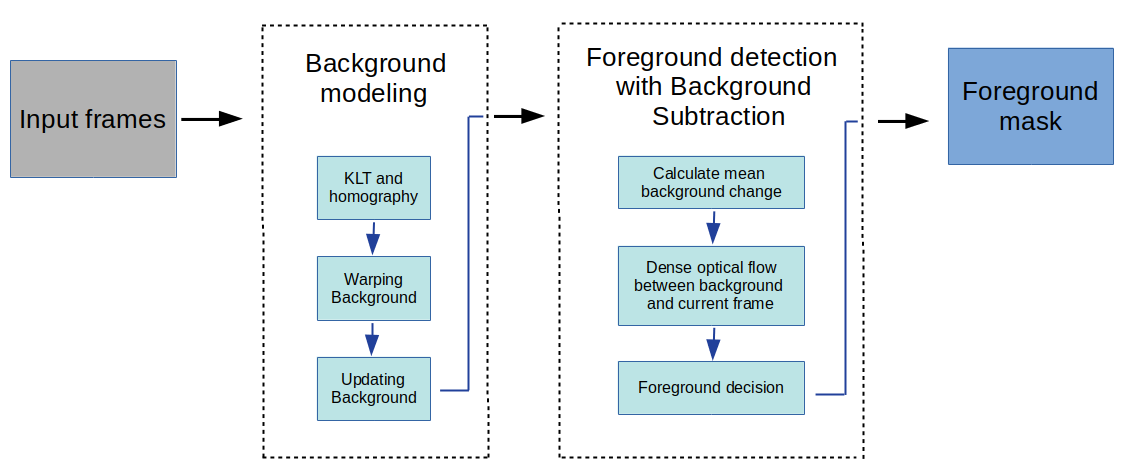}
\caption{Framework of proposed method}
\label{fig:framework}
\end{figure}

The general flow diagram of the proposed method is shown in the Figure \ref{fig:framework}. In the first step, grid-based points are selected in previous frame and these points are tracked by KLT method for current frame. Grid based points selection is used instead of the determining the key points to speed up the process. After finding the locations of the selected points in the current frame with the KLT method, the homograpy matrix(H) expressing the camera movement is calculated with the RANSAC method. At this stage, it is assumed that the background movement is much more dominant in the image. While building the background model, RGB frames are converted to HSV color space and only S and V channels are used. In the background model, mean value ($\mu$(i)) and age information of each pixel are calculated. Initially, the first frame is assigned to the $\mu$ matrix in the background model, and the age value is increased by 1 for each frame starting from 0.

\begin{figure}[b]
\centering
\includegraphics[width=\textwidth]{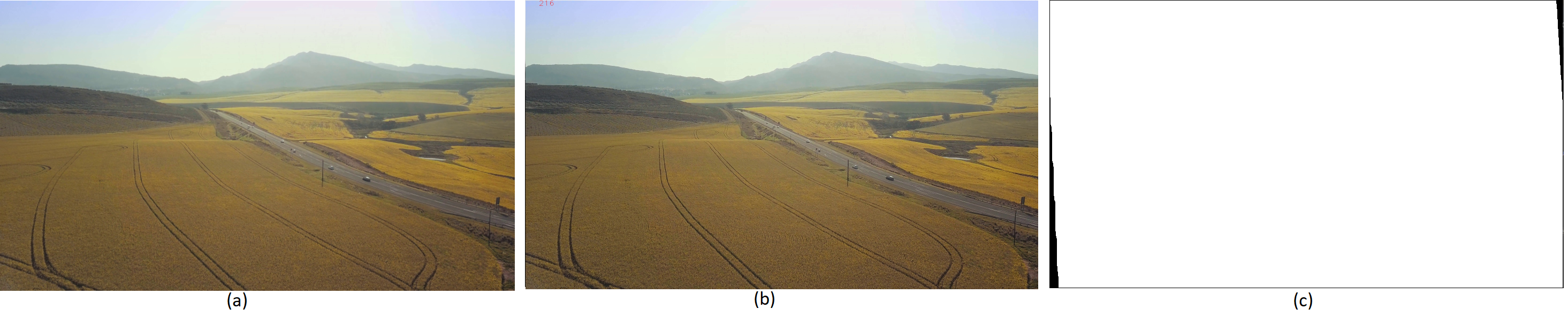}
\caption{(a) A frame at the time of t-1 ($I_{t-1}$) (b) $I_t$ (c) age mask}
\label{fig:BGandAgeMask}
\end{figure}

With the obtained $H$ matrix, the background model at time of \emph{t-1} is warped to the current frame as given in equation \ref{eq:homography}. Thus, the pixels in the background and current frame are aligned. But, there are possible registration errors, as an exact match is not possible. Using the $H$ matrix, the age value is assigned to 0 for the pixels that have just appeared in the current frame. Accordingly, if the camera is moving, a very small part of the current frame does not overlap with the previous frame. In Figure \ref{fig:BGandAgeMask}, example frames and age mask are shown. The black pixels in the age mask indicate the pixels with an age less than $T_{age}$. The overlapping pixels between the background and the current frame are shown in the Figure \ref{fig:BGandAgeMask} (c) as white, and only these pixels are evaluated in the foreground detection. In the background model, the $\mu$ value of each pixel is calculated as shown in the equation \ref{eq:mean} after the warping process. In the equations, $I$ represents a frame while $i$ represents a pixel in a frame. Learning rates ($\alpha$) of each pixel is determined with the \emph{age} value of each pixel.

\begin{equation}
B_t = H_{t-1} \: B_{t-1}
\label{eq:homography}
\end{equation}

\begin{equation}
\alpha(i) = \frac{1} {age(i)}
\label{eq:learning}
\end{equation}

\begin{equation}
\mu_t(i) = (1-\alpha_t(i))\:\mu_{t-1}(i) + \alpha(x)\ I_t(i)
\label{eq:mean}
\end{equation}

When performing foreground detection, only pixels with an age larger than threshold $T_{age}$ were taken into account. The $I_t$ is then subtracted from the $\mu$ matrix. In the background subtraction process, the neighborhood difference is calculated as given in the equation \ref{eq:diff} by considering the neighbours of each pixel in a 3x3 neighborhood area.

\begin{equation}
D(i) = min(\sum_{k=i-1}^{k=i+1} I_t(k)-\mu_{t-1}(i))
\label{eq:diff}
\end{equation}

Although using the neighborhood difference increases the computational load, it reduces the false detections caused by registration errors. The pixel differences calculated separately on the S and V channels are combined in a single matrix. Accordingly, the largest value for a pixel in the S and V matrix is chosen for each pixel. In this calculated difference image, a threshold value ($T$) could be used for foreground detection. But instead of using a fixed $T$, an adaptive threshold ($T_a$) is determined, considering that there may be more registration errors when the camera moves fast. To handle this, the threshold value is adjusted according to the background motion. Consider that $A_{BG}$ value represents the average value of pixel displacements according to the grid based selected points, $T_a$ is found as in equation \ref{eq:adaptiveThresh}:

\begin{equation}
T_a = T * \lambda_1 * exp( A_{BG} * \lambda_2 )
\label{eq:adaptiveThresh}
\end{equation}

\begin{figure}[t]
\centering
\includegraphics[width=\textwidth]{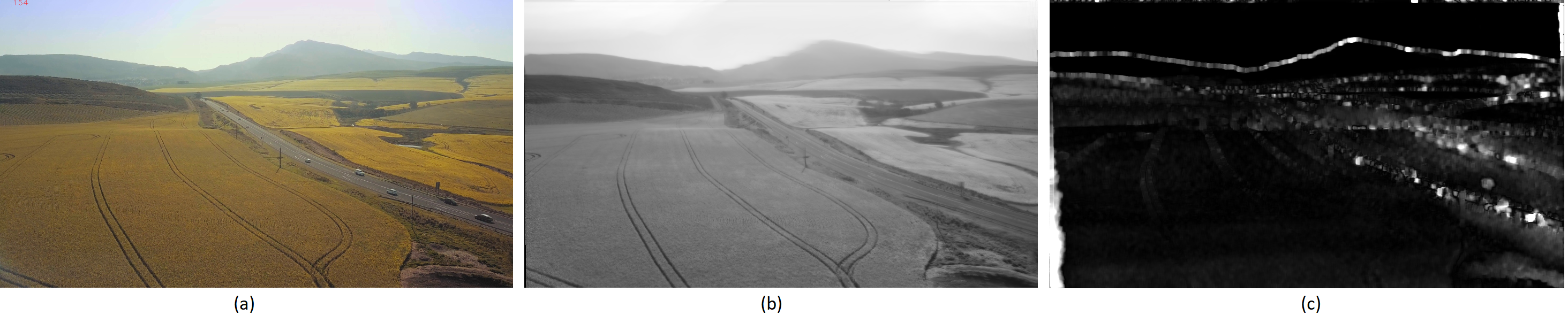}
\caption{(a) $I_t$ (b) Background model $\mu_t$ image (c) Magnitude of dense optical flow between $I$ and $\mu$}
\label{fig:magnitude}
\end{figure}

To improve the effect of the adaptive thresholding in background subtraction, dense optical flow is also calculated between $I_t$ and $\mu_t$ by Farneback \cite{farneback2003two} method, and optical flow vectors are also considered in the background subtraction process. For this, the magnitude matrix is calculated from the optical flow vectors and a weight coefficient is assigned to the pixels with high magnitude values as shown in equation \ref{eq:normalize}. Then, $D_w$ matrix is obtained by multipliying $D$ matrix and $Mag_n$. A sample $\mu$ image from the dataset and a magnitude image obtained with dense optical flow are shown in the Figure \ref{fig:magnitude}. Note that magnitude values are multiplied with a fixed number to be able to visualize it better. In Figure \ref{fig:magnitude}, it is seen that optical flow vectors are larger in the regions of the cars but it is also larger in the sharp edge such as the mountain. The error of Farneback method also reduces the effect of the using weights in background subtraction step.

\begin{equation}
Mag_{n} = norm(Mag) \:, where Mag(i) > T_{mag}
\label{eq:normalize}
\end{equation}

\begin{equation}
D^w(i) = Mag_{n}(i) \: D(i)  
\label{eq:normalizedDiff}
\end{equation}

In last step of background subtraction process, foreground mask is obtained by using $T_a$ and $D_w(i)$ as shown in the equation \ref{eq:fgMask}.

\begin{equation}
M^t(i) = \begin{cases} 
1 & D_w(i) > T_a \\
0 & otherwise
\end{cases}
\label{eq:fgMask}
\end{equation}

\begin{figure}[h]
\centering
\includegraphics[width=.7\textwidth]{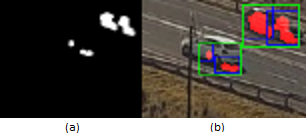}
\caption{ (a) Foreground mask (b) Extracted bounding boxes from foreground and bounding box combining}
\label{fig:regions}
\end{figure}

Finally, bounding boxes enclosing moving targets are extracted from foreground masks. For this, morphological opening process is applied to the foreground masks to suppress the noise, and bounding boxes covering areas with an area larger than 5 pixels are extracted. Subsequently, each bounding box is enlarged with a fixed value, and a larger bounding box enclosing the intersected boxes is obtained by combining the intersected boxes. In Figure \ref{fig:regions}, an example is shown to extract and combine the bounding boxes. Figure \ref{fig:regions} (b) shows detected moving pixels as red, blue rectangles represent extracted bounding box from foreground mask, and red rectangles represent the final bounding box after merging blue rectangles. The purpose of this process is to combine the broken foreground regions. Performance comparisons are made by comparing the finally found bounding boxes with labeled bounding boxes.

\section{Experimental Results}
\label{sec:results}

For the experiments, the proposed method was implemented by using C++ and OpenCV with CUDA support. The age value is set to 30 as maximum to keep minimum learning rate while $T_{age}$ is set to 5 for minimum age. $T$=40 value is used for foreground detection, and decaying parameters $\lambda_1$ and $\lambda_2$ for $T_a$ is 0.005 and 0.25 respectively. $T_{mag}$=5 value is used to decide if optical flow magnitude is enough large. Proposed method was experimented with eight image sequences in prepared PESMOD dataset that is publicly available in \url{http://github.com/mribrahim/PESMOD}

\subsection{Performance Comparison}
\label{sec:performance}

\begin{figure}[h]
\centering
\includegraphics[width=.7\textwidth]{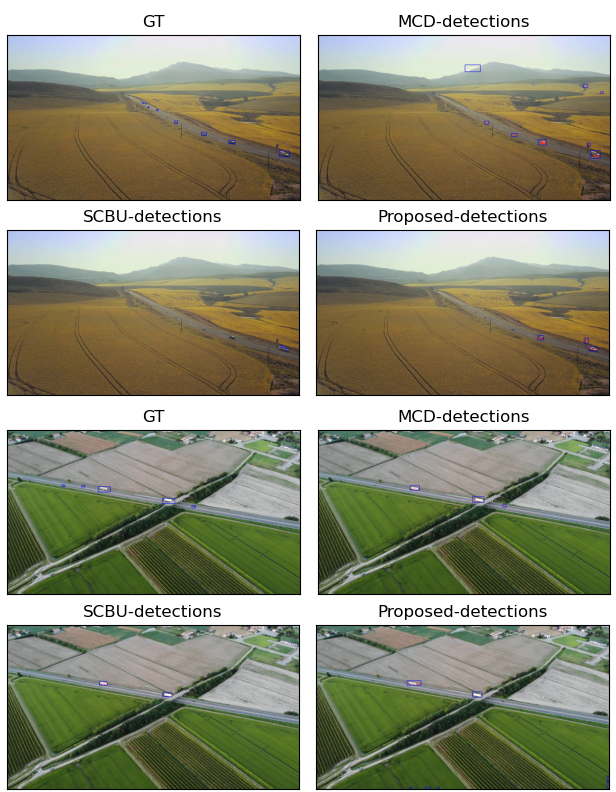}
\caption{ Detected bounding boxes for moving objects from \emph{Elliot-road} and \emph{Marian} sequences}
\label{fig:results-1}
\end{figure}

\begin{figure}[h]
\centering
\includegraphics[width=.7\textwidth]{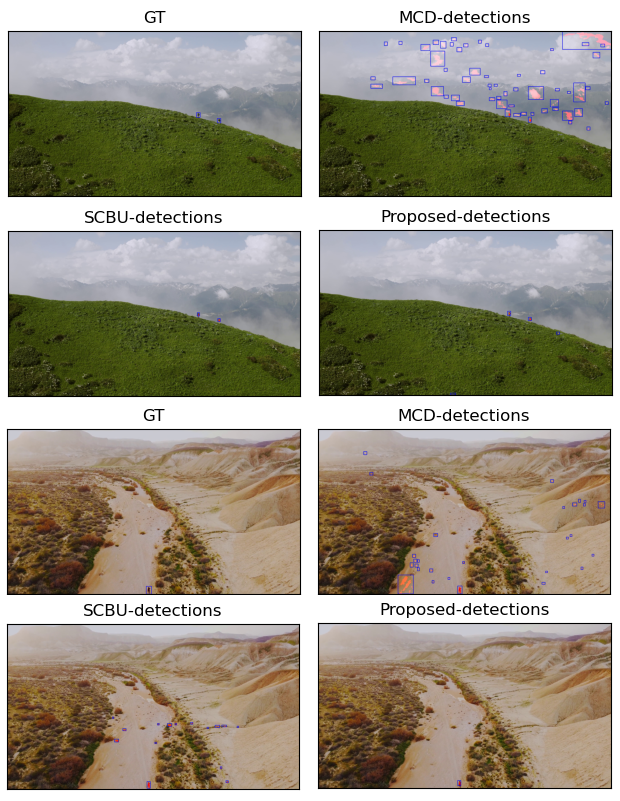}
\caption{ Detected bounding boxes for moving objects from \emph{Shuraev-trekking} and \emph{Zaborski} sequences}
\label{fig:results-2}
\end{figure}

Qualitative results on some key frames from each sequence are given in Figure \ref{fig:results-1} and \ref{fig:results-2}. They show foreground masks and extracted bounding boxes. In addition to qualitative comparison, precision ($Pr$), Recall ($R$), F-score ($F_1$) and overlapping ratio ($O_r$) for correctly detected moving targets are used for qualitative evaluation. In the equation \ref{eq:metrics}, $FP$ refers to wrongly detected boxes, $TP$ refers to true detections, and $FN$ refers to ground truth boxes that is missed by the method. In the experiments, it is seen that it is so hard to be able to find all pixels of a moving object. Therefore, in case of ground truth bounding box ($R_{GT}$) and detected bounding box ($R_{DET}$) intersect by 20 percent or more, it is evaluated as $TP$. $FP$ means that detected bounding box has no overlap even at 20 percent. Also another metric is defined to express the ratio of $R_{DET}$ and $R_{T}$ for all $TP$ detection as shown in equation \ref{eq:metricOR}. Higher $R_{DET}$ means that, it has found more parts of the moving object. The quantitative performance comparison is shown for each sequence in Table \ref{tab:performances}. Obtained average values for each metric is also given in Table \ref{tab:performanceAverage}.

\begin{equation}
Pr = \frac{TP}{TP+FP} ,  R = \frac{TP}{TP+FN} , F_1 = \frac{2 * Pr * R}{Pr + }
\label{eq:metrics}
\end{equation}

\begin{equation}
O_r = \frac{area(R_{GT} \& R_{DET})}{area(R_{GT})} 
\label{eq:metricOR}
\end{equation}

\begin{table}
\centering
\caption{Performance comparison of MCD, SCBU and proposed method}
\begin{tabular}{|l|l|l|l|l|l|l|l|l|} 
\hline
\multicolumn{1}{|c|}{\begin{tabular}[c]{@{}c@{}}\textbf{Sequence }\\\textbf{ name} \end{tabular}} & \multicolumn{1}{c|}{\textbf{} } & \multicolumn{1}{c|}{\textbf{MCD} } & \multicolumn{1}{c|}{\textbf{SCBU} } & \multicolumn{1}{c|}{\textbf{Proposed} } & \multicolumn{1}{c|}{\begin{tabular}[c]{@{}c@{}}\textbf{Sequence }\\\textbf{ name} \end{tabular}} & \multicolumn{1}{c|}{\textbf{MCD} } & \multicolumn{1}{c|}{\textbf{SCBU} } & \multicolumn{1}{c|}{\textbf{Proposed} }  \\ 
\hline
\multirow{4}{*}{\textit{Elliot-road} }                                                            & $O_r$                             & \textbf{0.8706}                    & 0.4141                              & 0.5939                                  & \multirow{4}{*}{\textit{Marian} }                                                                & \textbf{0.8462}                    & 0.5904                              & 0.7360                                   \\ 
\cline{2-5}\cline{7-9}
                                                                                                  & $Pr$                              & \textbf{0.3645}                    & 0.1458                              & 0.2987                                  &                                                                                                  & \textbf{0.9895}                    & 0.6706                              & 0.9231                                   \\ 
\cline{2-5}\cline{7-9}
                                                                                                  & $R$                               & \textbf{0.4022}                    & 0.163                               & 0.1010                                  &                                                                                                  & \textbf{0.5754}                    & 0.3971                              & 0.4395                                   \\ 
\cline{2-5}\cline{7-9}
                                                                                                  & $F_1$                              & \textbf{0.3824}                    & 0.0294                              & 0.1510                                  &                                                                                                  & \textbf{0.7276}                    & 0.4988                              & 0.5955                                   \\ 
\hline
\multirow{4}{*}{\textit{Miksanskiy} }                                                             & $O_r$                             & 0.8495                             & 0.5413                              & \textbf{0.8690}                         & \multirow{4}{*}{\textit{Grisha-snow} }                                                           & \textbf{0.7692}                    & 0.2871                              & 0.7184                                   \\ 
\cline{2-5}\cline{7-9}
                                                                                                  & $Pr$                              & \textbf{0.9417}                    & 0.7236                              & 0.2264                                  &                                                                                                  & 0.8984                             & 0.1696                              & \textbf{0.9645}                          \\ 
\cline{2-5}\cline{7-9}
                                                                                                  & $R$                               & \textbf{0.9417}                    & 0.7619                              & 0.8783                                  &                                                                                                  & \textbf{0.2000}                    & 0.0895                              & 0.1192                                   \\ 
\cline{2-5}\cline{7-9}
                                                                                                  & $F_1$                              & \textbf{0.9417}                    & 0.7422                              & 0.3600                                  &                                                                                                  & \textbf{0.3271}                    & 0.1172                              & 0.2123                                   \\ 
\hline
\multirow{4}{*}{\textit{Shuraev-trekking} }                                                       & $O_r$                             & \textbf{0.8391}                    & 0.4847                              & 0.6198                                  & \multirow{4}{*}{\textit{Zaborski} }                                                              & 0.6292                             & 0.5997                              & \textbf{0.6655}                          \\ 
\cline{2-5}\cline{7-9}
                                                                                                  & $Pr$                              & 0.0514                             & \textbf{0.6913}                     & 0.4217                                  &                                                                                                  & 0.2059                             & 0.4268                              & \textbf{0.7713}                          \\ 
\cline{2-5}\cline{7-9}
                                                                                                  & $R$                               & \textbf{0.9373}                    & 0.8625                              & 0.8909                                  &                                                                                                  & 0.6777                             & 0.8607                              & \textbf{0.9123}                          \\ 
\cline{2-5}\cline{7-9}
                                                                                                  & $F_1$                              & 0.0975                             & \textbf{0.7675}                     & 0.5724                                  &                                                                                                  & 0.3159                             & 0.5706                              & \textbf{0.8359}                          \\ 
\hline
\multirow{4}{*}{\textit{Welton} }                                                                 & $O_r$                             & \textbf{0.8991}                    & 0.5800                              & 0.7420                                  & \multirow{4}{*}{\textit{Wolfgang} }                                                              & 0.5026                             & 0.4620                              & \textbf{0.7601}                          \\ 
\cline{2-5}\cline{7-9}
                                                                                                  & $Pr$                              & 0.0728                             & \textbf{0.4203}                     & 0.2901                                  &                                                                                                  & \textbf{0.6747}                    & 0.1671                              & 0.4662                                   \\ 
\cline{2-5}\cline{7-9}
                                                                                                  & $R$                               & \textbf{0.6740}                    & 0.4322                              & 0.3919                                  &                                                                                                  & 0.0777                             & 0.1131                              & \textbf{0.3233}                          \\ 
\cline{2-5}\cline{7-9}
                                                                                                  & $F_1$                              & 0.1314                             & \textbf{0.4262}                     & 0.3334                                  &                                                                                                  & 0.1393                             & 0.1349                              & \textbf{0.3818}                          \\
\hline
\end{tabular}
\label{tab:performances}
\end{table}

\begin{table}[h]
\centering
\caption{Average performance metrics obtained with MCD, SCBU and proposed method}
\begin{tabular}{|l|l|l|l|}
\hline
\multicolumn{1}{|c|}{Average} & \multicolumn{1}{c|}{\textbf{MCD}} & \multicolumn{1}{c|}{\textbf{SCBU}} & \multicolumn{1}{c|}{\textbf{Proposed}} \\ \hline
$O_r$                           & \textbf{0.7756}                   & 0.4949                             & 0.6755                                 \\ \hline
$Pr$                            & 0.5428                            & 0.4268                             & \textbf{0.5452}                        \\ \hline
$R$                             & \textbf{0.5607}                   & 0.4416                             & 0.5070                                 \\ \hline
$F_1$                            & 0.3828                            & 0.4108                    & \textbf{0.4302}                                 \\ \hline
\end{tabular}
\label{tab:performanceAverage}
\end{table}

Figure \ref{fig:results-1} and \ref{fig:results-2} show ground truth bounding boxes and detected boxes for each method in four sequences: \emph{Elliot-road}, \emph{Marian}, \emph{Shuraev-trekking} and \emph{Zaborski}. In the experiments, it is seen that especially in case of fast camera movements, MCD produces much FP according to other methods as shown in first and third rows in Figure \ref{fig:results-2}. This causes to decrease in MCD performance even if MCD has best $Pr$ or $R$ for more sequences. For example; in \emph{Shuraev-trekking} and \emph{Welton} sequences, MCD has best $R$ but so less $Pr$ compared to the other method as seen in Table\ref{tab:performances}.

Also, it is observed that MCD method is more successful to detect smaller objects against SCBU and proposed method. In Figure \ref{fig:results-1}, only MCD can detect the small car, the others can only detect the big trucks. Table \ref{tab:performanceAverage} shows the average values for each metric and it shows that MCD has best $O_r$ and $R$, while proposed method have best $F_1$ and $Pr$. SCBU has best performance in $F_1$ after proposed method.  Best $O_r$ for MCD shows that it can detect the more part of a moving object compared to other methods. Proposed method generally performs poorly for zooming frames such as \emph{Miksanskiy} and \emph{Welton} sequences. It also has lowest $R$ value on frames with very small objects such as \emph{Elliot-road} and \emph{Grisha-snow}. Relatively simple background updating technique used in the proposed method may cause this issue.

\section{Conclusion}
\label{sec:conclusion}

In this paper, a motion detection method is proposed for moving cameras and the dataset called PESMOD is shared publicly. Proposed method applies motion compensation with KLT algorithm and neighborhood difference with the contribution of dense optical flow. We have compared the proposed method with state-of-the-art methods on various moving camera videos. The results show that even if the methods in the literature have more complex background model updating methods, they perform poorly for some special cases. However, proposed method shows a reasonable performance and it runs runs at about 17 fps for 1920x1080 resolution images.

\bibliographystyle{unsrt}

\begin{thebibliography}{10}

\bibitem{collins2005open}
Robert Collins, Xuhui Zhou, and Seng~Keat Teh.
\newblock An open source tracking testbed and evaluation web site.
\newblock In {\em IEEE International Workshop on Performance Evaluation of
  Tracking and Surveillance}, volume~2, page~35, 2005.

\bibitem{wang2014cdnet}
Yi~Wang, Pierre-Marc Jodoin, Fatih Porikli, Janusz Konrad, Yannick Benezeth,
  and Prakash Ishwar.
\newblock Cdnet 2014: An expanded change detection benchmark dataset.
\newblock In {\em Proceedings of the IEEE conference on computer vision and
  pattern recognition workshops}, pages 387--394, 2014.

\bibitem{collins2000system}
Robert~T Collins, Alan~J Lipton, Takeo Kanade, Hironobu Fujiyoshi, David
  Duggins, Yanghai Tsin, David Tolliver, Nobuyoshi Enomoto, Osamu Hasegawa,
  Peter Burt, et~al.
\newblock A system for video surveillance and monitoring.
\newblock {\em VSAM final report}, 2000(1-68):1, 2000.

\bibitem{zhao2011study}
Lixing Zhao, Qikai Tong, and Hongrui Wang.
\newblock Study on moving-object-detection arithmetic based on w4 theory.
\newblock In {\em 2011 2nd International Conference on Artificial Intelligence,
  Management Science and Electronic Commerce (AIMSEC)}, pages 4387--4390. IEEE,
  2011.

\bibitem{bouwmans2014traditional}
T~Bouwmans, B~Hofer-lin, F~Porikli, and A~Vacavant.
\newblock Traditional approaches in background modeling for video surveillance.
\newblock {\em Handbook Background Modeling and Foreground Detection for Video
  Surveillance, Taylor and Francis Group, T. Bouwmans, B. Hoferlin, F. Porikli,
  A. Vacavant}, 2014.

\bibitem{moo2013detection}
Kwang Moo~Yi, Kimin Yun, Soo Wan~Kim, Hyung Jin~Chang, and Jin Young~Choi.
\newblock Detection of moving objects with non-stationary cameras in 5.8 ms:
  Bringing motion detection to your mobile device.
\newblock In {\em Proceedings of the IEEE Conference on Computer Vision and
  Pattern Recognition Workshops}, pages 27--34, 2013.

\bibitem{zivkovic2004improved}
Zoran Zivkovic.
\newblock Improved adaptive gaussian mixture model for background subtraction.
\newblock In {\em Proceedings of the 17th International Conference on Pattern
  Recognition, 2004. ICPR 2004.}, volume~2, pages 28--31. IEEE, 2004.

\bibitem{zivkovic2006efficient}
Zoran Zivkovic and Ferdinand Van Der~Heijden.
\newblock Efficient adaptive density estimation per image pixel for the task of
  background subtraction.
\newblock {\em Pattern recognition letters}, 27(7):773--780, 2006.

\bibitem{allebosch2015efic}
Gianni Allebosch, Francis Deboeverie, Peter Veelaert, and Wilfried Philips.
\newblock Efic: edge based foreground background segmentation and interior
  classification for dynamic camera viewpoints.
\newblock In {\em International conference on advanced concepts for intelligent
  vision systems}, pages 130--141. Springer, 2015.

\bibitem{de2017wisardrp}
Massimo De~Gregorio and Maurizio Giordano.
\newblock Wisardrp for change detection in video sequences.
\newblock In {\em ESANN}, 2017.

\bibitem{huang2018optical}
Junjie Huang, Wei Zou, Jiagang Zhu, and Zheng Zhu.
\newblock Optical flow based real-time moving object detection in unconstrained
  scenes.
\newblock {\em arXiv preprint arXiv:1807.04890}, 2018.

\bibitem{ilg2017flownet}
Eddy Ilg, Nikolaus Mayer, Tonmoy Saikia, Margret Keuper, Alexey Dosovitskiy,
  and Thomas Brox.
\newblock Flownet 2.0: Evolution of optical flow estimation with deep networks.
\newblock In {\em Proceedings of the IEEE conference on computer vision and
  pattern recognition}, pages 2462--2470, 2017.

\bibitem{yun2017scene}
Kimin Yun, Jongin Lim, and Jin~Young Choi.
\newblock Scene conditional background update for moving object detection in a
  moving camera.
\newblock {\em Pattern Recognition Letters}, 88:57--63, 2017.

\bibitem{tomasi1991detection}
Carlo Tomasi and Takeo Kanade.
\newblock Detection and tracking of point.
\newblock Technical report, features. Technical Report CMU-CS-91-132, Carnegie,
  Mellon University, 1991.

\bibitem{fischler1981random}
Martin~A Fischler and Robert~C Bolles.
\newblock Random sample consensus: a paradigm for model fitting with
  applications to image analysis and automated cartography.
\newblock {\em Communications of the ACM}, 24(6):381--395, 1981.

\bibitem{farneback2003two}
Gunnar Farneb{\"a}ck.
\newblock Two-frame motion estimation based on polynomial expansion.
\newblock In {\em Scandinavian conference on Image analysis}, pages 363--370.
  Springer, 2003.

\end{thebibliography}

\end{document}